\documentclass[journal]{IEEEtran}
\IEEEoverridecommandlockouts
\usepackage{amsmath,amsfonts}
\usepackage{algorithmic}
\usepackage{algorithm}
\usepackage{array}
\usepackage[caption=false,font=normalsize,labelfont=sf,textfont=sf]{subfig}
\usepackage{textcomp}
\usepackage{stfloats}
\usepackage{url}
\usepackage{verbatim}
\usepackage{graphicx}
\usepackage{hyperref}
\usepackage{cite}
\usepackage{multirow}
\usepackage{bigstrut}
\usepackage{booktabs}
\usepackage{xcolor}
\usepackage{fancyhdr} 
\usepackage{orcidlink}
\usepackage[normalem]{ulem}
\def\BibTeX{{\rm B\kern-.05em{\sc i\kern-.025em b}\kern-.08em
    T\kern-.1667em\lower.7ex\hbox{E}\kern-.125emX}}
\definecolor{darggreen}{rgb}{0.1,0.8,0.3}
\ifCLASSINFOpdf
\else
   \usepackage[dvips]{graphicx}
\fi
\usepackage{url}

\usepackage{graphicx}

\begin{document}

\title{Multiple Object Tracking in Video SAR: A Benchmark and Tracking Baseline}

\author{Haoxiang Chen\orcidlink{0009-0005-2387-6616}, Wei Zhao\orcidlink{0000-0002-6060-1022}, Rufei Zhang, Nannan Li, and Dongjin Li
\thanks{\it{(Corresponding author: Wei Zhao.)}}
\thanks{Haoxiang Chen and Wei Zhao are with the School of Electronic and Information Engineering, Beihang University, Beijing 100191, China (e-mail: hxchen1230@buaa.edu.cn; zhaowei203@buaa.edu.cn).}
\thanks{Rufei Zhang, Nannan Li amd Dongjin Li are with the Beijing Institute of Control and Electronics Technology, Beijing, 100038, China (e-mail: zrf\_lucky@sina.com; beyondlnn@163.com; dongjindddj@163.com).}}

\markboth{IEEE GEOSCIENCE AND REMOTE SENSING LETTERS}
{Shell \MakeLowercase{\textit{et al.}}: Bare Demo of IEEEtran.cls for IEEE Journals}
\maketitle

\begin{abstract}
In the context of multi-object tracking using video synthetic aperture radar (Video SAR), Doppler shifts induced by target motion result in artifacts that are easily mistaken for shadows caused by static occlusions.
Moreover, appearance changes of the target caused by Doppler mismatch may lead to association failures and disrupt trajectory continuity.
A major limitation in this field is the lack of public benchmark datasets for standardized algorithm evaluation.
To address the above challenges, we collected and annotated 45 video SAR sequences containing moving targets, and named the Video SAR MOT Benchmark (VSMB).
Specifically, to mitigate the effects of trailing and defocusing in moving targets, we introduce a line feature enhancement mechanism that emphasizes the positive role of motion shadows and reduces false alarms induced by static occlusions.
In addition, to mitigate the adverse effects of target appearance variations, we propose a motion-aware clue discarding mechanism that substantially improves tracking robustness in Video SAR.
The proposed model achieves state-of-the-art performance on the VSMB, and the dataset and model are released at \href{https://github.com/softwarePupil/VSMB}{https://github.com/softwarePupil/VSMB}.
\end{abstract}

\IEEEpeerreviewmaketitle
\section{Introduction}
\IEEEPARstart{V}{ideo} SAR, with its spatio-temporal continuity and all-weather imaging, enables stable acquisition of dynamic information in complex environments.
It is widely used in border patrol, maritime monitoring, and downstream tasks such as behavior recognition and event prediction \cite{VSARM}.

\begin{figure}[!t]
   \setlength{\abovecaptionskip}{0pt}   
   \setlength{\belowcaptionskip}{-2pt}   
   \centering
   \includegraphics[width=0.85\columnwidth]{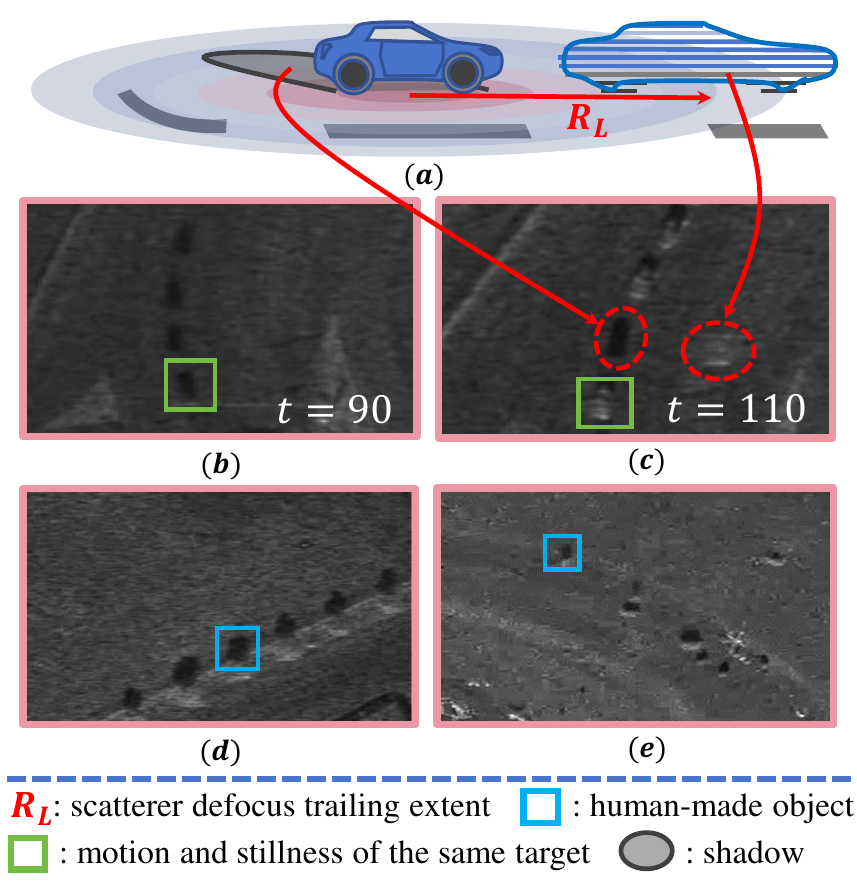}
   \caption{Challenges of MOT in Video SAR. (a) Doppler mismatch of moving targets. (b)(c) Appearance changes under different motion states. (d)(e) Shadows caused by human-made object.}
   \label{fig_1}
\end{figure}

Detection and tracking in video SAR still have the following challenges:
1) Doppler mismatch enhances sidelobe energy \cite{TMI}, resulting in trailing artifacts of scattering points along the azimuth direction and frequent feature transitions between shadows and scatterers, which lead to association failures and trajectory fragmentation, as shown in Fig. \ref{fig_1} (b) and (c).
2) Shadows caused by radar beam occlusion are difficult to distinguish from those of moving targets, often leading to false alarms, as shown in Fig. \ref{fig_1} (d) and (e).
3) The lack of open-source benchmarks impedes the development of Video SAR.

Several recent studies \cite{RFBMFC,SE-SA-AAN,Dual-Mode Framework,GNN-JFL,GraphMTT,JTMT,MFJD} have begun to explore target tracking in video SAR.
Studies \cite{RFBMFC,SE-SA-AAN,Dual-Mode Framework} incorporate spatio-temporal clues, motion shadow features, and environmental background constraints into the detection process.
Studies \cite{GNN-JFL,GraphMTT} integrate shadow appearance and motion clues of moving targets to design more robust graph neural network-based association models.
Studies \cite{JTMT,MFJD} aim to improve detector robustness by employing cross-frame energy detection methods.
However, these methods primarily aim to suppress trailing shadows and defocus effects, while overlooking the positive effect of these effects in reducing false alarms.

To address the above challenges, this letter releases the open-source VSMB dataset and proposes a MOT method, named Video SAR MOT Track (VSMT).
Firstly, we collect high-quality video SAR data containing moving targets from open sources and perform frame-by-frame annotation using DarkLabel.
Secondly, to facilitate evaluation, we construct a benchmark by integrating the VSMB with representative Joint Detection Tracking (JDT) and Detection-Based Tracking (DBT) models from recent MOT frameworks.
Inspired by \cite{TMI} and \cite{RST}, we observe that the trailing and defocusing effects of moving targets play a crucial role in distinguishing between shadows cast by stationary objects and moving targets.
Therefore, Line Feature Focusing Module (LFFM) and Line Feature Assigner (LFA) are introduced to help proposals perceive surrounding line features and reduce false alarms from human-made objects.
Finally, Motion-awareness Association (MaA) is introduced to address trajectory fragmentation caused by target appearance changes.
MaA dynamically adjusts the weight of appearance similarity in the association cost matrix according to the motion state of the target.
Extensive experiments demonstrate that VSMT achieves state-of-the-art performance in both detection and tracking on the VSMB.
The main contributions of this letter are as follows:
\begin{enumerate}{}{}
   \item{To the best of our knowledge, VSMB is the first publicly available dataset for video SAR target tracking, providing a unified and valuable benchmark for future research.}
   \item{The LFFM and LFA are proposed to reduce false alarms and trajectory errors by leveraging the trailing and defocusing characteristics of moving targets.}
   \item{The MaA module dynamically enables the use of appearance similarity in the association process based on motion state awareness, effectively reducing trajectory errors.}
  \end{enumerate} 

\section{The Proposed Dataset}
\label{section_2}
\subsection{Data Collection and Annotation}
\label{subsection_1}
\begin{table}[t]
  \setlength{\abovecaptionskip}{-2pt}
  \centering
  \caption{Overview of VSMB Data Sources}
  \renewcommand{\arraystretch}{0.8} 
  \begin{tabular}{lllll}
  \toprule
  Video                         & Enterprise/        & \multirow{2}{*}{Platform}   & \multirow{2}{*}{Resolution}  & Observed            \\ 
  Number                        & Organization       &                             &                              & Area                 \\ \cmidrule(l){1-5}
  $1\sim 4$                     & ICEYE              & ICEYE-SAR                   &  0.5-15m	                    & Airport	              \\
  $5$                           & AIRSAT             & Haishao-1                   &	2m                          &	Port                   \\
  $6\sim10$                     & ICEYE              & ICEYE-SAR                   &	0.5-15m                     &	Mine                    \\
  $11\sim27$                    & SNL                & Airborne SAR                &	0.1-0.3m                    &	Transport                \\ 
  $28\sim40$                    & ICEYE              & ICEYE-SAR                   &	0.5-15m                     &	Mine             	        \\ 
  $41$                          & SS                 & SmartSat-X1                 &	1m                          &	Port              	       \\
  $42\sim45$                    & ICEYE              & ICEYE-SAR                   &	0.5-15m                     &	Port              	        \\ \bottomrule
  \end{tabular} 
  \label{table_sources}
\end{table}
The SAR MOT dataset VSMB \footnote{\url{https://github.com/softwarePupil/VSMB}} constructed in this study includes video clips from various representative scenarios that are freely and publicly released by ICEYE, SNL\cite{SNL}, AIRSAT, and SS (Smart Satellite), as summarized in Table \ref{table_sources}.
The collected video SAR data cover representative scenes such as airports, ports, urban areas, and mining regions.
Following the splitting strategy in DOTA-devkit \cite{DOTA}, we partition large images into smaller tiles of size $1024\times 1024$.
In addition, the training dataset is expanded by employing a data augmentation strategy based on video rewinding.
To ensure annotation quality, DarkLabel is used for frame-by-frame labeling in strict accordance with a MOT annotation standard.

\subsection{Dataset Overview}
\label{subsection_2}
\begin{table}[t]
  \setlength{\abovecaptionskip}{-2pt}
  \centering
  \caption{Overview of VSMB Statistics}
  \renewcommand{\arraystretch}{0.8} 
  \begin{tabular}{llllll}
  \toprule
  \multirow{2}{*}{Category}     & \multirow{2}{*}{Instance}        & Total                       & Small                        & Medium             & Large            \\ 
                                &                                  & Trajector                   & $(<16^2)$                    & $(<32^2)$          & $(<64^2)$         \\ \cmidrule(l){1-6}
  Car                           & 19986                            & 306                         &  8.67\%	                    & 35.09\%	           & 99.24\%            \\
  Ship                          & 4362                             & 41                          &	0.05\%                      &	30.67\%            & 64.92\%             \\
  Airplane                      & 1118                             & 8                           &	0.09\%                      &	0.54\%             & 23.08\%              \\ \cmidrule(l){1-6}
  Total                         & 25466                            & 355                         &	6.81\%                      &	32.82\%            & 90.02\%               \\ \bottomrule
  \end{tabular} 
  \label{table_statistics}
\end{table}
The constructed VSM dataset includes three target categories: cars, ships, and airplanes.
Table \ref{table_statistics} presents a statistical summary of the dataset, which consists of 45 videos and 4,394 images, covering 25,466 target instances and 355 trajectories.

As shown in Fig. \ref{fig_1}, the target experiences Doppler shift, and the resulting image sequence is often accompanied by significant defocusing effects.
In addition, shadows cast by moving targets are often visually similar to those caused by static man-made occlusions, which can easily lead to false alarms in detection and tracking.

\section{Baseline Model}
\label{section_3}
Current MOT methods can be broadly categorized into three groups: JDT (e.g., \cite{FairMOT,CenterTrack}), DBT (e.g., \cite{ByteTrack,BoTSORT,OCSORT,StrongSORT,UAVMOT}), and Implicit Association Tracking (IAT) (e.g., \cite{MOTRv2,MOTIP}).
JDT and IAT achieve joint optimization of detection and association via multi-task parallel learning.
However, in the video SAR scenarios of the VSMB, the DBT approach achieves the best performance, as shown in Table \ref{table_VSMT}.
JDT and IAT adopt multi-task optimization to integrate detection and association, which inherently introduces conflicts in feature sharing.
If a slight misalignment exists between tasks, it is amplified in small target scenarios, leading to a significant reduction in detection accuracy.

In summary, considering the characteristics of VSMB, we adopt the detection-based tracking framework Dino \cite{DINO} for detection and ByteTrack \cite{ByteTrack} for tracking (Dino-Byte) as the baseline model.

\section{Proposed Method}
\label{section_3}
\subsection{Overview}
\label{subsection_1}
\begin{figure*}[!t]
  \setlength{\abovecaptionskip}{0pt}   
  \setlength{\belowcaptionskip}{-2pt}   
  \centering
  \includegraphics[width=0.9\textwidth]{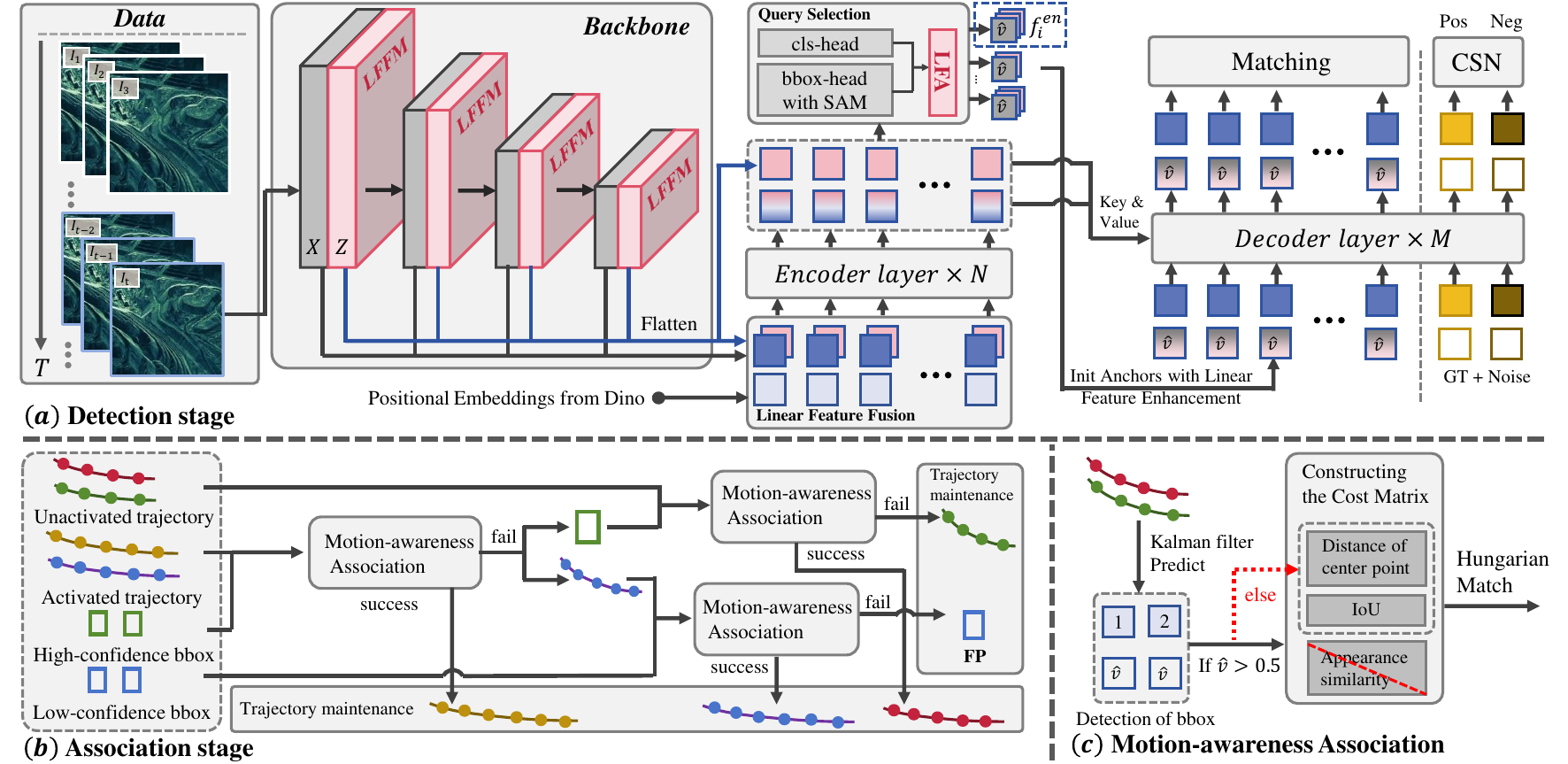}
  \caption{Schematic diagram of VSMT. (a) Detection stage of VSMT with Dino as a baseline model, LFFM and LFA are highlighted in \textcolor{red}{Red}. (b) Association stage of VSMT with ByteTrack as a baseline model. (c) Schematic flow diagram of MaA.}
  \label{fig_2}
\end{figure*}

The MOT task in SAR video aims to maintain the identity (ID) of each target over time and generate a corresponding set of trajectories ${{\cal T}_*} = \{ {T_n}\} _{n = 1}^N$, where $N$ denotes the total number of trajectories.
As shown in Fig. \ref{fig_2}, given a SAR video sequence ${\cal I}  = \{ {I_t}\} _{t = 1}^T$, a window of $W$ consecutive frames is first collected and concatenated along the channel dimension.
The resulting tensor is then fed into the detector Dino to obtain the detection result ${D_t} = \{ b_t^m\} _{m = 1}^{{M_t}}$, where $M_t$ denotes the total number of observed targets at time $t$.
Each target observation $b_t^m = [t,{\rm{id}},x,y,w,h,s,c,v]$ consists of a bounding box $[x,y,w,h]$ , and attributes $[s,c,v]$ denote the confidence score, category label, and motion awareness level, respectively.

As shown in Fig. \ref{fig_2}, VSMT integrates the LFFM module into each scale feature to enhance the multi-scale representations after feature spreading via the Linear Feature Fusion (LFF) mechanism.
Meanwhile, the output of LFFM is passed to LFA, where selective linear feature enhancement is applied to the proposals from the Dino encoder output.
The object association process and MaA logic of VSMT are shown in Fig. \ref{fig_2} (b) and (c).

\subsection{Line Feature Focusing Module}
\label{subsection_2}
The study \cite{RST} found that the contextual features of targets are highly discriminative and can effectively mitigate the interference caused by defocusing and trailing effects.
Therefore, VSMT adopts the DETR-Base \cite{DINO} model due to its strong capability in context modeling.
In addition, VSMT further leverages line features caused by trailing and defocusing to guide the model in distinguishing shadows of stationary objects from those of moving targets.

Although the backbone performs well in extracting point-like targets \cite{CBNet}, it is less effective for elongated linear structures.
Inspired by \cite{Radon} and \cite{CCLDet}, LFFM converts linear features from trailing and defocusing into pointwise representations that are more easily perceived by the network.

\begin{figure}[t]
  \setlength{\abovecaptionskip}{-2pt}   
  \setlength{\belowcaptionskip}{-2pt}   
  \centering
  \includegraphics[width=0.9\columnwidth]{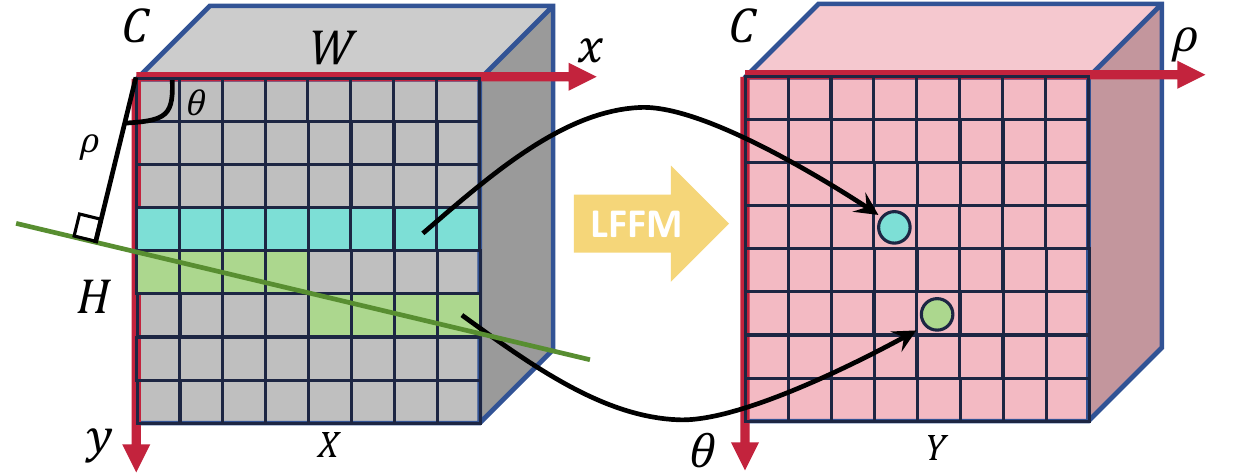}
  \caption{LFFM transforms $X$ in the spatial domain to $Y$ in the Radom domain.}
  \label{fig_Radon}
\end{figure}
As shown in Fig. \ref{fig_2}, LFFM first applies a discrete Radon transform to the input feature map $X \in {\mathbb{R}^{{H_F} \times {W_F} \times C}}$ to obtain the feature distribution $Y \in {\mathbb{R} ^{{\Theta _F} \times {P_F} \times C}}$ in the Radon domain.
The Radon transform accumulates the response values along a line segment $L$ in the feature map and projects them to a single point $({\hat \theta _F},{\hat \rho _F})$ in the Radon domain:
\begin{align}
  {\bf{Y}}({\hat \theta _F},{\hat \rho _F}) &= \sum\limits_{l \in L} {\bf{X}} (l), \\
  {\hat \theta _F} = \left\lfloor {\frac{{{\theta _F}}}{{\Delta {\theta _{{\rm{Rdn}}}}}}} \right\rfloor &,{\hat \rho _F} = \left\lfloor {\frac{{{\rho _F}}}{{\Delta {\rho _{{\rm{Rdn}}}}}}} \right\rfloor,
\end{align}
where ${\theta _F}$ and ${\rho _F}$ correspond to successive positions along the line $l \in L$ in the Radon domain.
And $0 \le {\theta _F} < \pi$, $-\sqrt{H_F^2 + W_F^2} / 2 \leq \rho_F \leq \sqrt{H_F^2 + W_F^2} / 2$.
The angular resolution $\Delta {\theta _{{\rm{Rdn}}}}$ and radial resolution $\Delta {\rho _{{\rm{Rdn}}}}$ are calculated as follows:
\begin{align}
  \Delta {\theta _{{\rm{Rdn}}}} &= \frac{\pi }{{{\Theta _F}}}, \Delta {\rho _{{\rm{Rdn}}}} = \frac{{\sqrt {{H_F}^2 + {W_F}^2} }}{{{P_F}}},
\end{align}
To obtain a linear feature enhancement map, LFFM reconstructs the feature representation using an inverse projection accumulation mechanism (Inverse Radon Transform).
Each nonzero point $({\hat \theta _F},{\hat \rho _F})$ in $Y$ corresponds to a straight line in the image domain, defined as follows:
\begin{align}
  {L_{{{\hat \theta }_F},{{\hat \rho }_F}}} &= \left\{ {({x_F},{y_F}){\mkern 1mu} |{\mkern 1mu} \rho  = x_F\cos \theta  + y_F\sin \theta } \right\} \\
  \theta  &= {\hat \theta _F} \cdot \Delta {\theta _{{\rm{Rdn}}}}, \rho  = {\hat \rho _F} \cdot \Delta {\rho _{{\rm{Rdn}}}}
\end{align}
where $({x_F},{y_F})$ is a point on the spatial feature map $X$. The line feature intensity map ${{\bf{A}}_{{\rm{soft}}}}$ is computed as follows:
\begin{align}
  \mathbf{A}(x_F, y_F, c) = \sum_{\hat{\theta}_F, \hat{\rho}_F} 
  & \mathbb{I} \left( \mathbf{Y}(\hat{\theta}_F, \hat{\rho}_F, c) \ge \tau \right) 
  \mathbf{Y}(\hat{\theta}_F, \hat{\rho}_F, c) \nonumber \\
  &  \cdot \mathbb{I} \left( (x_F, y_F) \in L_{\hat{\theta}_F, \hat{\rho}_F} \right), \\
  {{\bf{A}}_{{\rm{soft}}}}({x_F},{y_F},c) = &\frac{{\exp ({\bf{A}}({x_F},{y_F},c))}}{{\sum\limits_{{{x'}_F},{{y'}_F}} {\exp } ({\bf{A}}({{x'}_F},{{y'}_F},c))}},
\end{align}
where $\mathbb{I} \left( \cdot \right)$ is an indicator function, and  controls the filtering of noise.
Ultimately, ${{\bf{A}}_{{\rm{soft}}}}$ represents a line feature response map. 
It exhibits strong activations in dense regions of short line segments, while showing weak responses in sparse regions.
\begin{align}
  [{{\bf{W}}_X},{{\bf{W}}_A}] &= \sigma ({\rm{Con}}{{\rm{v}}_{1 \times 1}}([{\bf{X}},{{\bf{A}}_{{\rm{soft}}}}])), \\
  {\bf{Z}} &= ({{\bf{W}}_X} + {\bf{I}}) \odot {\bf{X}} + {{\bf{W}}_A} \odot {{\bf{A}}_{{\rm{soft}}}},
\end{align}
where $\left[\cdot \right]$ denotes the feature concatenation operator, $\sigma$ is the sigmoid activation function, and ${\bf{I}}$ is the identity matrix.

\subsection{Line Feature Assigner}
\label{subsection_3}
Existing MOT models \cite{VSARM,RFBMFC,SE-SA-AAN,Dual-Mode Framework,GNN-JFL,GraphMTT,JTMT,MFJD,RST} mainly rely on point features for proposal prediction.
Due to the insufficient exploitation of distinctive line features in SAR imaging, the above models tend to confuse static and dynamic shadows during the decoding phase.
To address the above issue, SVMT employs the LFA.
LFA is based on the line feature intensity map generated by LFFM, which is selectively processed and integrated into the proposal prediction stage of the Dino encoder.
The study \cite{TMI} indicates that the Doppler shift of a scattering point can be expressed as:
\begin{align}
  f_d(x, y) = \frac{2}{\lambda} \left( \frac{dx_r}{dt} - \frac{dx_a}{dt} \right), \label{equation_tmi_1}
\end{align}
where $\lambda $ is the radar wavelength, ${dx_r}/{dt}$ denotes the relative velocity between the radar and the target center, and ${dx_a}/{dt}$ represents the velocity of the scattering point relative to the target center.
As the ${dx_r}/{dt}$ increases, the Doppler shift becomes more pronounced, and the distance between the scattering streak and the target body in the image increases.
We add a dimension to the bounding box head and supervise it using L1 loss, with the target being the spatial displacement of the ground truth box between consecutive frames:
\begin{align}
  v_t^i &= \frac{{\sqrt {{{(x_t^i - {\mathop{\rm CMC}\nolimits} (x_{t - 1}^i))}^2} + {{(y_t^i - {\mathop{\rm CMC}\nolimits} (y_{t - 1}^i))}^2}} }}{{\Delta f}}, \\
  \tilde v_t^i &= \frac{{v_t^i}}{{{{\max }_{t,j}}v_t^j}},
\end{align}
where $\rm CMC(\cdot )$ is derived from \cite{CMC}, and $(x_t^i, y_t^i)$ denotes the center position of the $i$-th sample in the video at time $t$.
LFA determines the matching radius range for each proposal based on its corresponding line feature intensity map ${{\bf{A}}_{{\rm{soft}}}}$.
Based on Equation (\ref{equation_tmi_1}), the velocity ${dx_r}/{dt}$ is incorporated into the radius calculation to adaptively adjust the matching range.
The feature matching and fusion process of LFA is described as follows:
\begin{align}
  {R_i} &= {R_{\min ,i}} + {\hat v_i}\left( {{R_{\max ,i}} - {R_{\min ,i}}} \right), \\
  {{\cal N}_i} &= \left\{ {({x_j},{y_j}){\mkern 1mu} |{\mkern 1mu} \sqrt {{{({x_j} - {x_i})}^2} + {{({y_j} - {y_i})}^2}}  \le {R_i}} \right\}, \\
  f_i^{{\rm{en}}} &= f_i^{{\rm{prop}}}  \nonumber \\
  &+ MLP(\frac{1}{{|{{\cal N}_i}|}}\sum\limits_{({x_j},{y_j}) \in {{\cal N}_i}} {{A_{{\rm{soft}}}}} [{x_j},{y_j},:]),
\end{align}
where ${R_{\min ,i}} = \max ({w_i},{h_i})$, ${R_{\max ,i}} = {\lambda _{\max }}\max ({x_i},W - {x_i},{y_i},H - {y_i})$.
$\left\{ {f_i^{pop}} \right\}_{i = 1}^K$ denotes the features of the top-K proposals from the Dino decoder.
The final augmented feature $f_i^{{\rm{en}}}$ combines the local contextual information from ${{A_{{\rm{soft}}}}}$ with the original representation of the proposal.

\section{EXPERIMENTS}
\label{section_4}
\subsection{Experimental Setup}
\label{subsection_D_3}
\begin{table*}[!t]
  \setlength{\abovecaptionskip}{-2pt}
  \centering
  \caption{Quantitative Results in VSMB. ↑ Indicates That Higher is Better,↓ Indicates That Lower is Better.The Best Results are 
  Shown in Bold \textcolor{red}{Red} and the Second Best Results are Shown in Bold Black. † Indicates a Graph-Based Models}
  \renewcommand{\arraystretch}{0.8} 
  \setlength{\tabcolsep}{6pt} 
  \begin{tabular}{c|c|c|ccccccccccc}
  \toprule
  Method                          & Backbone                   & DBT                              & MOTA$\uparrow$                           & IDSW$\downarrow$             & MT$\uparrow$                    & ML $\downarrow$                   & IDF1$\uparrow$                     & IDR$\uparrow$                  & IDP$\uparrow$                     & HOTA$\uparrow$                     & DetA$\uparrow$                  & AssA $\uparrow$       \\ \cmidrule(l){1-13}
  FairMOT \cite{FairMOT}          & X101                       &                                  & 37.1                                     & 224	                         & 133	                           & 137  	                           & 63.7                               & 64.0                           & 63.5    	                         & 50.0                              & 41.8                            & 60.8                   \\
  CenterTrack \cite{CenterTrack}  & X101                       &                                  & 48.0                                     & 450	                         & 116	                           & 81  	                             & 62.5                               & 57.2                           & 68.9    	                         & 50.7                              & 45.4                            & 57.6                   \\
  MOTRv2 \cite{MOTRv2}            & Swin-L                     &                                  & 52.2                                     & 357	                         & 107	                           & 89  	                             & 64.3                               & 54.1                           & 79.3   	                         & 51.1                              & 45.4                            & 58.2                    \\
  MOTIP \cite{MOTIP}              & Swin-L                     &                                  & 66.4                                     & 601	                         & 167	                           & 32	                               & 69.7                               & 64.8                           & 75.4   	                         & 56.7                              & 57.4                            & 57.0                     \\
  Yolov7-Bot \cite{BoTSORT}       & E-ELAN                     & $\checkmark$                     & 73.1                                     & 124	                         & 238	                           & 48	                               & 82.7                               & 78.3                           & 87.6                     	       & 67.7                              & 61.2                            & 74.9                      \\
  OC-SORT \cite{OCSORT}           & X101                       & $\checkmark$                     & 68.1                                     & 121	                         & 156                             & 77	                               & 79.6                               & 70.1                           & \bf{92.3}                    	   & 62.4                              & 55.6                            & 71.2                       \\
  StrongSORT \cite{StrongSORT}    & Swin-L                     & $\checkmark$                     & 75.7                                     & 113          	               & \bf{243}	                       & \bf{24}	                         & 83.6                               & 81.4                           & 85.9   	                         & \bf{68.8}                         & \bf{61.5}                       & \bf{76.9}                   \\
  Dino-Byte \cite{ByteTrack}      & Swin-L                     & $\checkmark$                     & 74.9                                     & \textcolor{red}{\bf{53}}	   & 219	                           & 47    	                           & 84.3                               & 76.8                           & \textcolor{red}{\bf{93.3}}   	   & 65.8                                & 60.3                            & 71.7                         \\ \cmidrule(l){1-13}
  VSMT (Ours)                     & Swin-S                     & $\checkmark$                     & \bf{77.3}                                & 109                          & \bf{243}                        & 29                                & \bf{85.8}                          & \bf{82.3}                      & 89.7   	                         & 62.7                                & 58.0                            & 72.8                          \\ 
  VSMT (Ours)                     & Swin-L                     & $\checkmark$                     & \textcolor{red}{\bf{78.2}}               & \bf{95}                      & \textcolor{red}{\bf{249}}       & \textcolor{red}{\bf{22}}          & \textcolor{red}{\bf{86.1}}         & \textcolor{red}{\bf{85.7}}     & 86.5       	                     & \textcolor{red}{\bf{70.3}}      	   & \textcolor{red}{\bf{64.6}}      & \textcolor{red}{\bf{78.0}}     \\ \bottomrule
  \end{tabular}
  \label{table_VSMT}
\end{table*}

\begin{table}[t]
  \setlength{\abovecaptionskip}{-2pt}
  \centering
  \caption{Ablation experiments of the VSMT in the VSMB. ↑ Indicates That Higher is Better, ↓ Indicates That Lower is Better}
  \renewcommand{\arraystretch}{0.8} 
  \begin{tabular}{ccc|cccc}
  \toprule
  LFFM                          &LFA                     &MaA                     & MOTA$\uparrow$          & HOTA$\uparrow$             & IDF1 $\uparrow$       & IDSW$\downarrow$        \\ \cmidrule(l){1-7}
                                &                        &                        & 75.2           	        & 66.4	                     & 82.1	                 & 193               \\
  $\checkmark$                  &                        &                        & 77.4                    &	67.8                       &	83.2                  & 167                \\
                                & $\checkmark$           &                        & 74.2                    &	68.1                       &	83.7                  & 178                 \\
                                &                        & $\checkmark$           & 75.4                    &	67.5                       &	84.3                  & 114 	                \\ 
  $\checkmark$                  & $\checkmark$           & $\checkmark$           & \bf{78.2}               &	\bf{70.3}                  &	\bf{86.1}             & \bf{95}               \\ \bottomrule
  \end{tabular} 
  \label{table_ablation}
\end{table}
VSMT is trained within the DBT framework using the AdamW optimizer for 36 epochs, with an initial learning rate of $2.5e-4$.
The input image size is fixed at $1024\times 1024$, and no additional data augmentation is applied unless necessary.
Under the premise of avoiding data leakage, the training and validation sets are divided in a ratio of $7:3$.

We use the MOT evaluation metrics, including: 
multiobject tracking accuracy (MOTA), false positives (FP), false negatives (FN), ID F1 score (IDF1), Higher Order Tracking Accuracy (HOTA), Detection Accuracy (DetA), and Association Accuracy(AssA).
To evaluate the effectiveness of the model, representative target trackers from recent years are selected for comparison.

\subsection{Ablation Study and Comparative Experiment}
\label{subsection_D_4}
\begin{figure}[t]
  \setlength{\abovecaptionskip}{-2pt}   
  \setlength{\belowcaptionskip}{-2pt}   
  \centering
  \includegraphics[width=0.9\columnwidth]{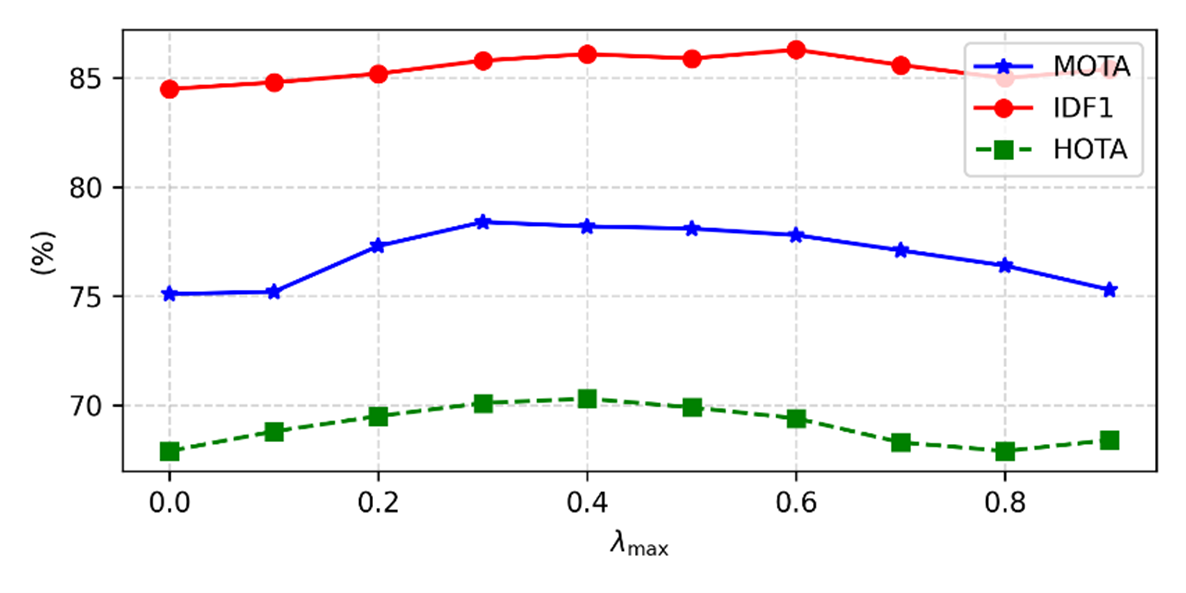}
  \caption{Experimental validation of LFA scaling factor ${\lambda _{\max }}$.}
  \label{fig_factor}
\end{figure}

As shown in Table \ref{table_ablation}, ablation experiments are conducted on the proposed modules LFFM, LFA, and MaA.
The addition of LFFM and LFA results in a significant increase in MOTA, indicating a mitigation of the FN and FP problems.
The introduction of MaA effectively mitigates the problem of trajectory fragmentation. 
Even when the target appearance changes during the transition from stationary to moving, VSMT is able to maintain ID continuity.
As shown in Fig. \ref{fig_factor}, we control the scaling factor ${\lambda _{\max }}$ for the maximum radius selected by LFA. 
Experimental results demonstrate that the model consistently achieves superior performance when ${\lambda _{\max }}$ is within the range $[0.2,0.6]$.

\begin{figure}[!t]
  \setlength{\abovecaptionskip}{0pt}   
  \setlength{\belowcaptionskip}{-2pt}   
  \centering
  \includegraphics[width=\columnwidth]{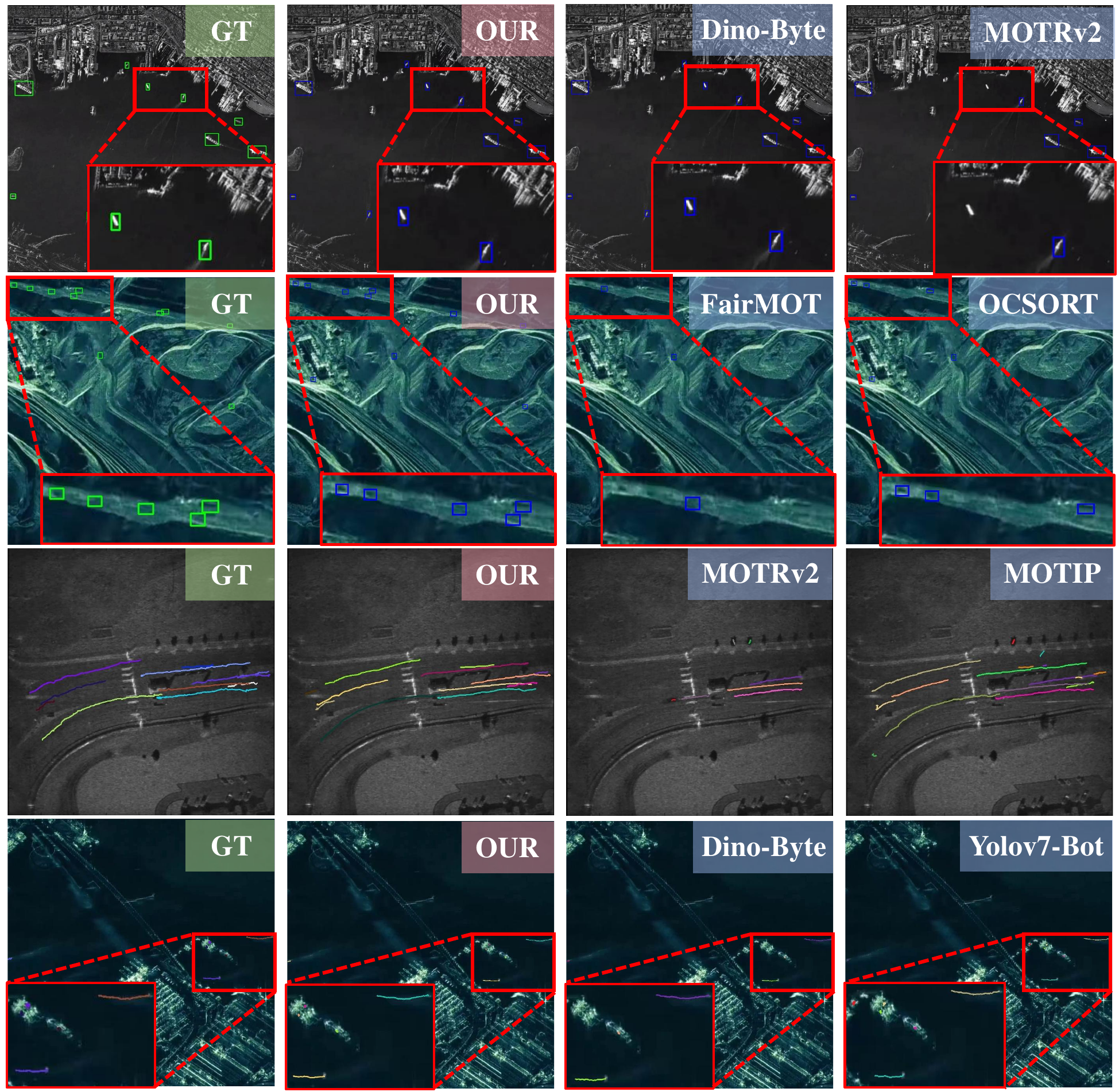}
  \caption{Examples of detections and trajectories. The \textcolor{darggreen}{green} box represents the GT, the \textcolor{blue}{blue} box signifies the output result, and identical target IDs share the same track color.}
  \label{fig_8}
\end{figure}

Extensive experiments are conducted to validate the performance of the baseline model Dino-Byte and demonstrate the superiority of the proposed VSMTrack.
As shown in Table \ref{table_VSMT}, VSMT achieves superior performance compared to recent state-of-the-art trackers in both tracking and detection tasks.
The effectiveness of VSMT in suppressing false alarms on human-made objects is further validated by the visualization results in Fig. \ref{fig_8}.

\subsection{Conclusion}
\label{subsection_D_6}
This letter presents key challenges and introduces the VSMB benchmark for video SAR target tracking, aiming to foster the development of models robust to appearance changes under strong interference.
To address the key challenges identified in the benchmark, this letter introduces VSMT.
VSMT employs line feature aggregation and enhancement to differentiate human-made shadows from target shadows, and motion-aware clues to modulate appearance similarity within the association cost matrix, effectively enhancing trajectory continuity.
Ultimately, the benchmark and baseline models are released as open source.
In the future, we will pay more attention to Doppler shift clues in Video SAR to supplement interference recognition.


\end{document}